\documentclass[english]{article}
\usepackage[T1]{fontenc}
\usepackage[latin9]{inputenc}
\usepackage{geometry}
\usepackage{fancyhdr}
\usepackage{color}
\usepackage{babel}
\usepackage{float}
\usepackage{graphicx}
\usepackage{setspace}
\usepackage[unicode=true]
 {hyperref}

\makeatletter

\providecommand{\tabularnewline}{\\}

\newcommand{\lyxaddress}[1]{
\par {\raggedright #1
\vspace{1.4em}
\noindent\par}
}

\pdfoutput=1
\AtBeginDocument{}
\date{}
\usepackage{multicol}

\makeatother

\begin{document}

\title{\textbf{\textcolor{black}{\large{}Speaker Fluency Level Classification
Using Machine Learning Techniques}}}

\author{Alan Preciado-Grijalva\thanks{Corresponding author: \protect\href{mailto:apreciado42@uabc.edu.mx}{apreciado42@uabc.edu.mx}},
and Ramon F. Brena\textit{\textcolor{black}{\small{} }}}

\maketitle
\vspace{-2.5em}

\lyxaddress{\begin{center}
\textit{\textcolor{black}{\small{}Grupo de Investigación en Sistemas
Inteligentes, Tecnológico de Monterrey,}}\textit{\small{} Monterrey,
NL, Mexico}
\par\end{center}}

\vspace{-3em}

\begin{abstract}

\vspace*{-.5em}

\textcolor{black}{\small{}Level assessment for foreign language students
is necessary for putting them in the right level group, furthermore,
interviewing students is a very time-consuming task, so we propose
to automate the evaluation of speaker fluency level by implementing
machine learning techniques. This work presents an audio processing
system capable of classifying the level of fluency of non-native English
speakers using five different machine learning models. As a first
step, we have built our own dataset, which consists of labeled audio
conversations in English between people ranging in different fluency
domains/classes (low, intermediate, high). We segment the audio conversations
into 5s non-overlapped audio clips to perform feature extraction on
them. We start by extracting Mel cepstral coefficients from the audios,
selecting 20 coefficients is an appropriate quantity for our data.
We thereafter extracted zero-crossing rate, root mean square energy
and spectral flux features, proving that this improves model performance.
Out of a total of 1424 audio segments, with 70\% training data and
30\% test data, one of our trained models (support vector machine)
achieved a classification accuracy of 94.39\%, whereas the other four
models passed an 89\% classification accuracy threshold.}{\small \par}

\end{abstract}

\begin{multicols}{2}

\section*{\textcolor{black}{\large{}\\1. Introduction}}

The development of artificial intelligence (AI) - powered applications
has been growing remarkably over the last decade \cite{key-1}\cite{key-2}.
With regards to language learning apps, there are currently several
software language companies that are employing AI techniques to improve
user engagement and learning experience. The main promise of AI-powered
language learning apps is that users will achieve basic proficiency
in a foreign language as they progress through their lessons within
a few months and with a small amount of time studying per day, all
being guided by AI. 

A slightly different language learning scenario is the one involving
two or more (known or unknown) persons who are actively looking for
tandem groups to improve their language skills. Currently, our group
at our university, the Tecnológico de Monterrey (ITESM), is working
on the construction of an AI-powered mobile app for language learning
called \textit{\textcolor{black}{Avalinguo}}. Avalinguo is an internet-based
system, and it merges \textcolor{black}{virtual reality with AI to
create ``digital classrooms'' in which people, each one with a corresponding
}\textit{\textcolor{black}{avatar}}\textcolor{black}{, can practice
a language \cite{key-3}. }

Avalinguo has many benefits such as 1) users are not attached to fixed
schedules, 2) it is portable and can be used anywhere (internet provided),
3) real-time real-person interaction, 4) user privacy is kept because
of the use of avatars, 5) it clusters users based on profiles (target
language, interests, etc.), 6) due to clustering, each login presents
new possible matches with other users (recommendation system), 7)
it contributes to a relaxed and casual participation by implementing
fluency monitoring and topic recommendation during conversations and
8) its cost is inferior to particular online courses. 

In this paper, we report work done related to point 7 by presenting
the advancements corresponding fluency monitoring during a conversation
between two or more persons. Our approach to this problem is based
on audio analysis, starting with audio feature extraction and afterwards
training machine learning (ML) models to perform classification of
audio segments provided labeled target classes. Previously, there
have been advancements in environmental sound classification \cite{key-4}
and real-time speech recognition based on neural networks \cite{key-5}.
In these cases, the audio sets have been environmental sounds (rain,
cars, birds, etc.) and recorded speech, music and noise sounds, respectively.
In our case, to approach the general problem of fluency level monitoring
of each individual during a conversation, we have first proceeded
to build our own audio set (\textit{\textcolor{black}{Avalinguo audio
set}}), the details of the audio set are presented in \textit{\textcolor{black}{section
2}}. Thereafter, we have split each conversation in 5s non-overlapped
segments, these segments have had some features extracted (mel coefficients
+ zero crossing rate + root-mean-square-energy + spectral flux). Later
on, the feature vectors are fed into a classification model to train
it and evaluate its performance using accuracy metrics. Our defined
fluency classes are three: low fluency, intermediate fluency, and
high fluency. We have compared five ML models, namely, multi-layer
perceptron (MLP), support vector machines (SVM), random forest (RF),
convolutional neural networks (CNN) and recurrent neural networks
(RNN). The workflow described previously is the standard ML approach
for the audio analysis of sound events \cite{key-6}.

The main question that we are trying to answer here is: \textit{\textcolor{black}{Given
a labeled balanced audio set fulfilling predefined fluency metrics,
can we construct a model capable of classifying the level of fluency
of an audio segment? }}

\textcolor{black}{If so, this would allow us to determine whether
a group in a conversation needs a recommended topic to keep it flowing.
Also, we would be able to tell if Person A has a lower fluency than
Person B and needs to be re-assigned to a lower fluency level group
(same for the inverse case).}

Our final classification results have achieved accuracies higher than
90\% (except for one model), being the highest of up to 94.39\% for
an SVM. As a first step, we have determined the appropriate number
of Mel coefficients (MFCCs) extracted to ensure high accuracies. Thereafter,
we've proved that adding features to the baseline MFCCs such as zero-crossing-rate
(ZCR), root-mean-squared-energy (RMSE) and spectral flux onset strength
envelope (SF) increased overall model performance.

\section*{\textcolor{black}{\large{}2. Avalinguo audio set}}

Having a clean-high quality data set is a must for any machine learning
project. The main challenge for a model's architecture is to be able
to grasp patterns among data so that it can be complex enough to perform
accurate predictions/classifications. This can be affected if mislabeled
or missing data is contained in the data set. On the other side, the
technical limitations of a model have to do with the computational
power available and the amount of labeled data required for appropriate
training.

Unfortunately for us, there are no\textit{\textcolor{black}{{} publicly}}
available data sets that fulfill our research purposes. Whilst the
community has been working extensively on the construction and support
of audio sets \cite{key-8}, it has not been possible for us to address
an audio set for audio-speech analysis composed of conversations by
non-native English speakers. There are indeed audio corpora of people
speaking English such as the UCSB and MUSAN {[}9-10{]}, but no audio
set of people who are actually learning the language (by this we mean
people who hesitate when speaking, people who take long pauses when
speaking, or also people who just speak too slowly). 

We are mostly interested in recordings of this kind because the Avalinguo
system will deal mainly with non-native English speakers who will
present the speaking characteristics mentioned before. Due to this
reason, we have decided to build our own audio set; \textit{\textcolor{black}{Avalinguo
audio set \cite{key-11}}}\textcolor{black}{. The Avalinguo audio
set is a collection of audio recordings of people whose language fluency
ranges from low to high. The audio recordings have the next common
characteristics:}
\begin{itemize}
\item Spontaneous (non-scripted) conversations
\item Random conversation topics chosen by speakers
\item Audios recorded with low-to-no background noise
\end{itemize}
The sources who provided the audio recordings are three: Friends/Family,
Language Center (ITESM), and Youtube \textcolor{black}{Audios. Each
audio recording comprises a conversation that lasts around 10 minutes
and has a wide range of topics from athletes speaking to physicists
talking about science to sisters talking about their daily activities.
All of these 10 minutes conversations were cut in 5s equally-sized
segments and were thereafter manually labeled and assigned into one
of the three fluency classes.} \textcolor{black}{Summarizing, the
audio set consists of 1420 (5s duration) non-overlapped audio segments
comprising three fluency classes (low, intermediate and high fluency).}
This is about 2 hours of recordings. The audios have sample rates
ranging from 22050 Hz to 48000 Hz, are mono and multi-channel and
were converted to MP3 format.

\textit{Figure 1} shows the class distribution of the audio set. The
intermediate class has a higher percentage because, technically, it
is easier to collect audios of people ranging in the intermediate
level rather than the low level. Despite this percentage differences,
we have kept all audio files to avoid reducing the size of the data
set. Note that this can be considered a small audio set (ca. 2 hours)
if it is compared with some public sets (such as Google AudioSet).

\begin{figure}[H]
\begin{centering}
\includegraphics[scale=0.5]{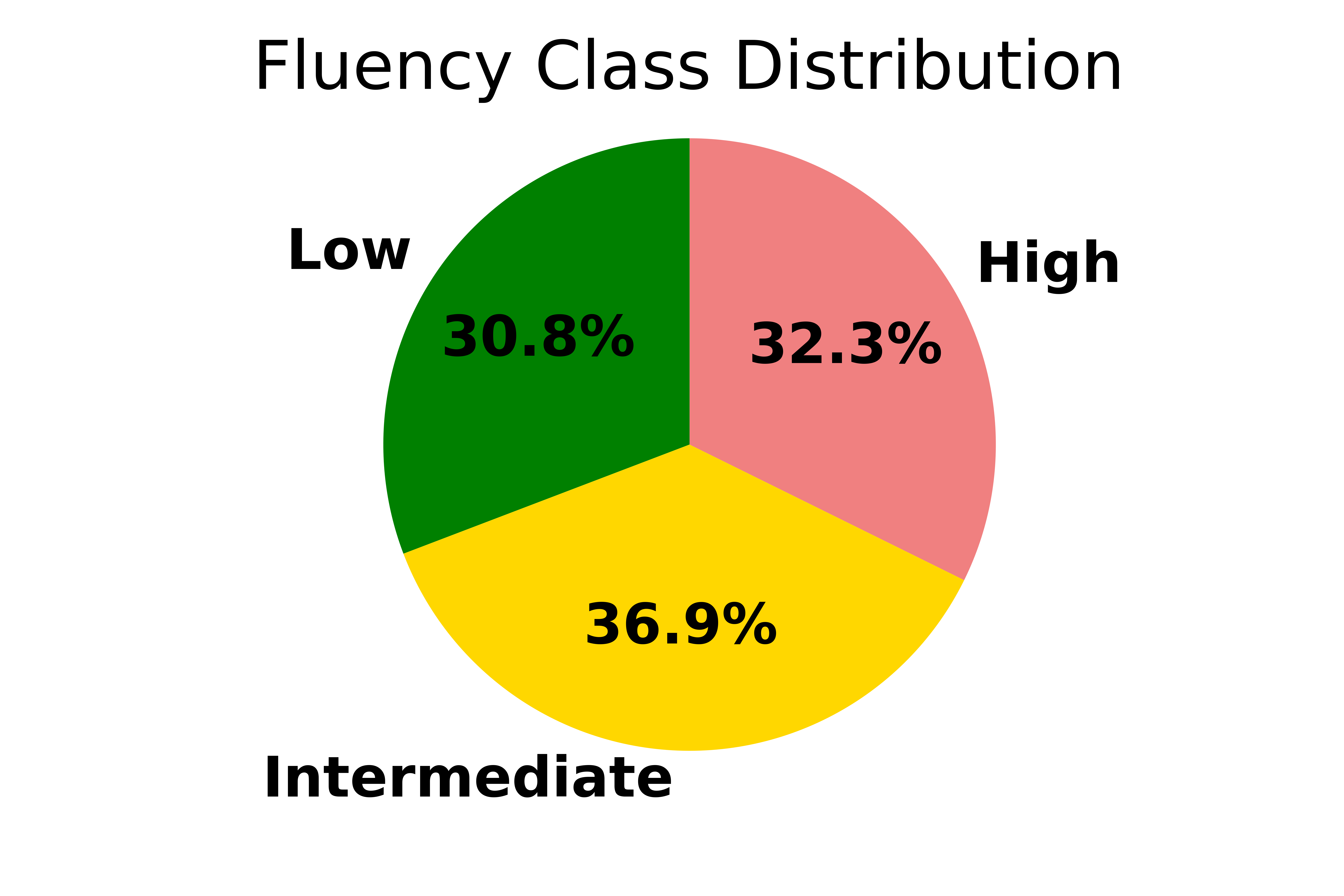}
\par\end{centering}
\centering{}\caption{Avalinguo audio set class distribution. The set contains a total of
118.65 minutes of recorded audio.}
\end{figure}

\subsection*{\textcolor{black}{\normalsize{}2.1 Fluency metrics}}

Previously, we mentioned that each audio segment was labeled manually.
In order to do this, we have defined baseline fluency levels definitions:

\textbf{Low 0:} Person uses very simple expressions and talks about
things in a basic way. Speaks with unnatural pauses. Needs the other
person to talk slowly to understand.

\textbf{Low 1}: Person can understand frequently used expressions
and give basic personal information. Person can talk about simple
things on familiar topics but still speaks with unnatural pauses. 

\textbf{Intermediate 2:} Can deal with common situations, for example,
traveling and restaurant ordering. Describes experiences and events
and is capable of giving reasons, opinions or plans. Can still make
some unnatural pauses. 

\begin{singlespace}
\textbf{Intermediate 3:} Feels comfortable in most situations. Can
interact spontaneously with native speakers but still makes prolonged
pauses or incorrect use of some words. People can understand the person
without putting too much effort.
\end{singlespace}

\textbf{High 4:} Can speak without unnatural pauses (no hesitation),
doesn't pause long to find expressions. Can use the language in a
flexible way for social, academic, and professional purposes.

\textbf{High 5:} Native-level speaker. Understands everything that
reads and hears. Understand humor and subtle differences.

There is no single-universal definition for fluency. Actually, each
language institution establishes a fluency metric for scoring based
on their internal parameters. In our case, to score speaker fluency,
we have taken the baseline definitions described above and have made
specific emphasis on the next points regarding the concept:

\textbf{1)} Our metric is mainly sound based. With \textquotedbl{}fluent\textquotedbl{}
meaning speaking without unnatural pauses.\textbf{ \\}

\textbf{2)} If there is hesitation (slowness or pauses) when speaking,
then that affects the fluency score of the speaker.\textbf{ \\}

\textbf{3)} There is a distinction between fluency and proficiency.
Meaning that fluency is someone able to feel comfortable, sound natural,
and with the ability to manipulate all the parts of a sentence at
will.

\subsection*{}

\section*{\textcolor{black}{\large{}3. Experimental Framework}}

Our experimental procedure is the standard ML approach for analysis
of sound events \cite{key-12}. Our system consists of three main
steps (see \textit{Figure 2}): Feature extraction, classification
and output of the predicted label with higher probability for individual
segments (frames). First, we perform feature extraction on each audio
frame. Thereafter, the feature vector is fed into a classifier which
outputs the most probable class based on previous training. 

\begin{figure}[H]
\begin{centering}
\includegraphics[scale=0.35]{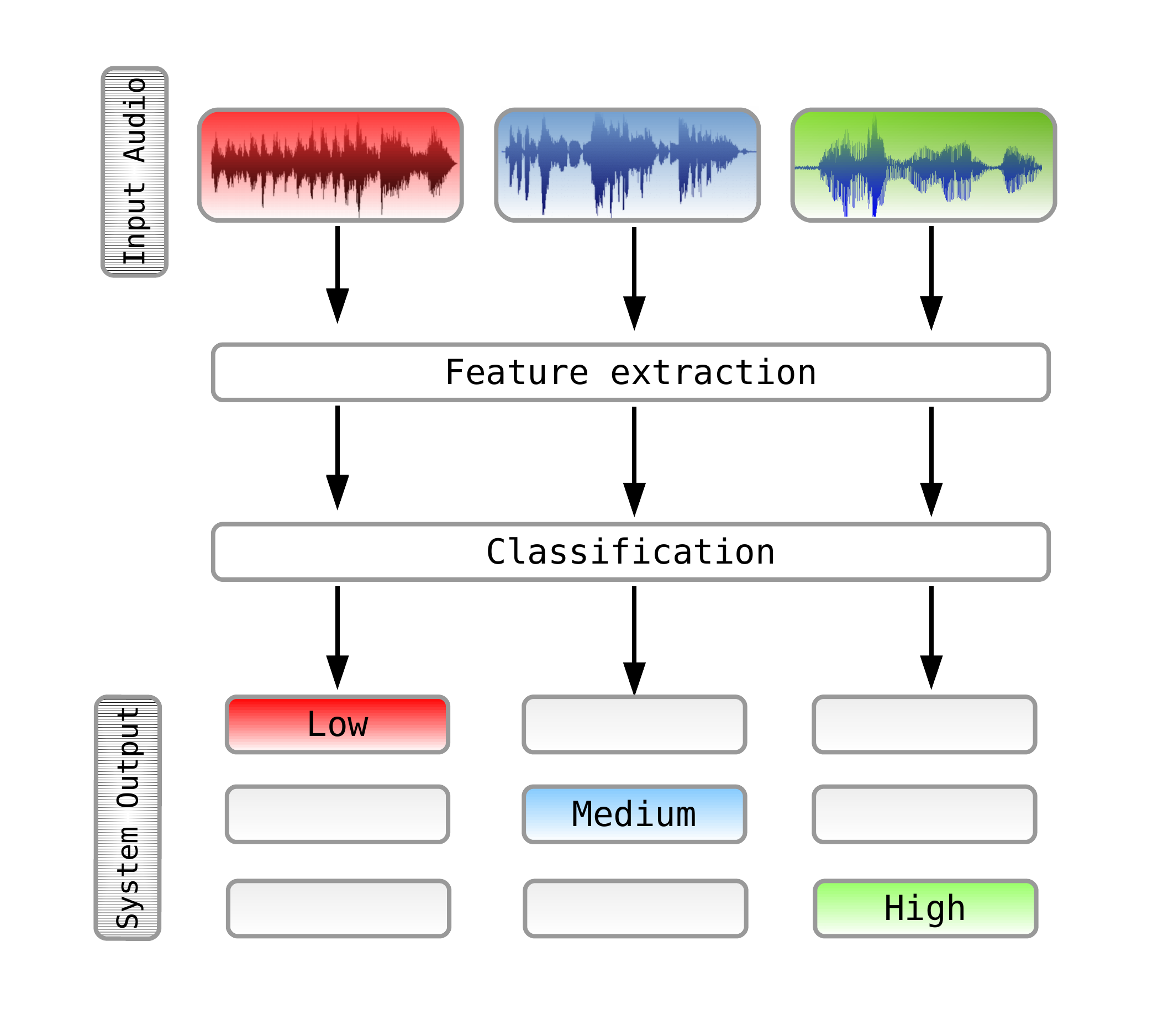}
\par\end{centering}
\centering{}\caption{Pipeline of an audio classification system.}
\end{figure}

\subsection*{\textcolor{black}{\normalsize{}3.1 Audio segmentation}}

Audio segmentation is one of the most important preprocessing steps
in most audio applications \cite{key-13}. Research groups often propose
novel segmentation frameworks in order to improve audio applications
such as speech recognition \cite{key-14}. In our case, our segmentation
method simply consists on cutting the audio files into 5s non-overlapped
segments and manually assign a frame to a single person (for example,
Frame 1 belongs to person A). If a frame contains more than one person
speaking, we assign it to the person that speaks more in it. After
this, we proceed to label frames according to their fluency level.
The segments were cut from the audios using a python module called
Pydub \footnote{\href{https://github.com/jiaaro/pydub}{Pydub: High level python interface for audio manipulation}}.

\begin{singlespace}
Due to the way we proceeded, we didn't work with overlapped segments;
since we are interested in differentiating persons among a conversation,
this approach would have had ended as non-overlapped segments but
with shorter time durations. The other approach that we explored was
using voice activity detection (VAD), this approach creates audio
segments when it detects a different voice or a pause during a conversation.
However, the VAD interface \cite{key-15} suppresses any possible
silence within the conversation, that is, it only creates segments
when people are speaking. Since we are interested in detecting possible
silences and pauses in each audio segment, we have discarded this
approach. 
\end{singlespace}
\begin{spacing}{0.7}

\subsection*{\textcolor{black}{\normalsize{}3.2 Feature Extraction}}
\end{spacing}

\begin{singlespace}
This is the step where features are extracted. We have written a python
script to perform feature extraction for the created audio segments
using a python package called Librosa \footnote{\href{https://librosa.github.io/librosa/}{LibROSA: python package for music and audio analysis.}}
(among many functionalities, Librosa is commonly used for feature
extraction, allowing to compute more than thirty audio features).
The audio frames $f_{0},...,f_{n}$, have thus their corresponding
feature vectors $p_{0},...,p_{n}$. 
\end{singlespace}

For the results presented in this paper, we have first varied the
number of MFCCs extracted, this with the purpose to estimate the appropriate
number of MFFCs to extract. Later on, we have added the ZCR, RMSE
and SF to determine if this improves model performance. 

Besides feature extraction, Librosa allows to plot audio spectrograms
using its \textit{display.specshow} function. For the sake of completeness,
we show some commonly spectrograms for a single audio frame (see\textit{
Figure 3}). 

\begin{figure}[H]
\begin{centering}
\includegraphics[scale=0.55]{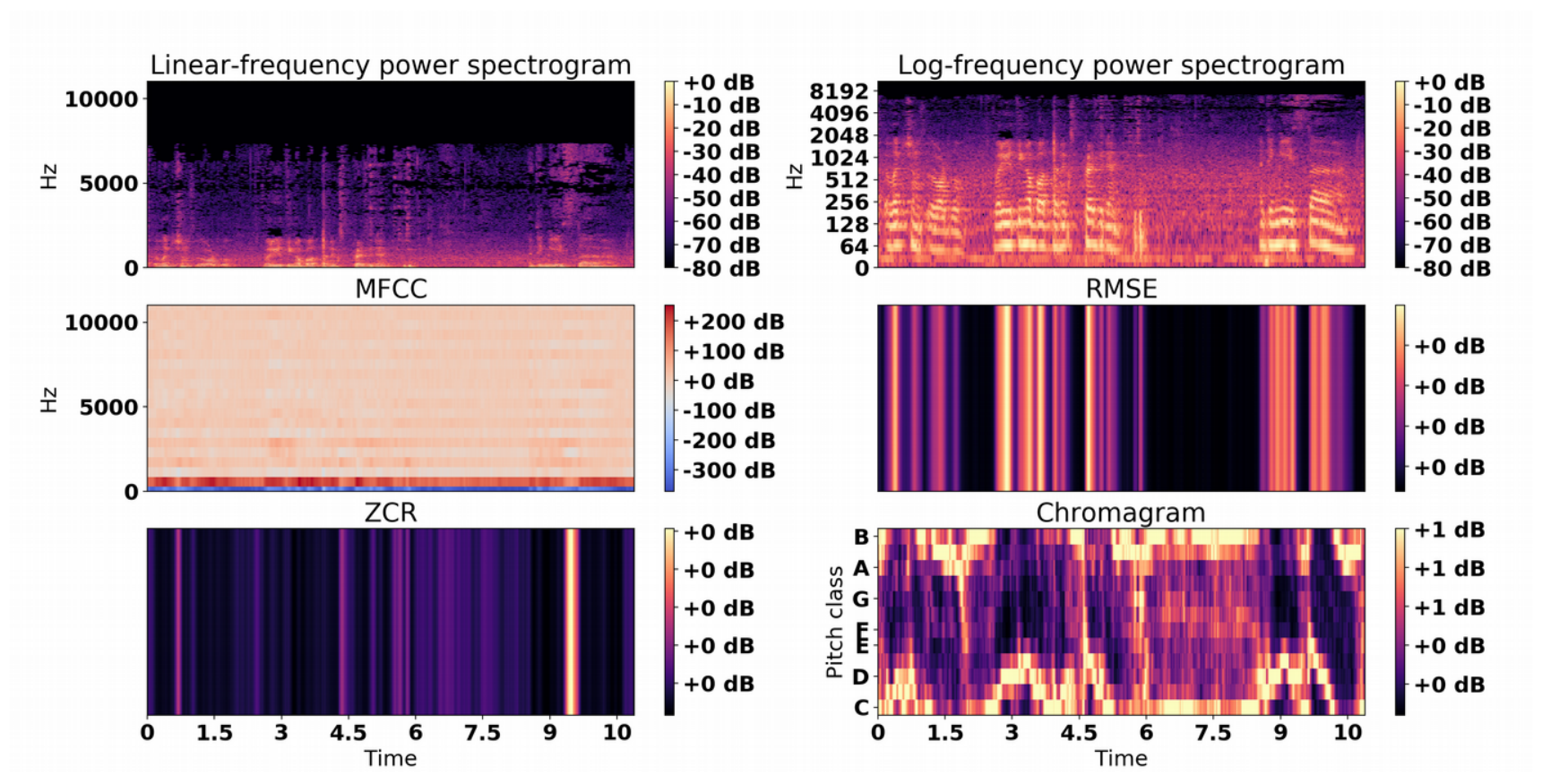}
\par\end{centering}
\centering{}\caption{Spectral feature plots (single frame).}
\end{figure}

\subsection*{\textcolor{black}{\normalsize{}3.3 Audio Classification}}

In this work we are interested in comparing different classification
frameworks, namely, multilayer perceptrons, convolutional neural networks,
recurrent neural networks, support vector machines and random forest.
The main goal here is to test each model classification accuracies
as we vary the number of features extracted.

The proposed multilayer perceptron architecture has two hidden layers
comprising 512 x 512 neurons followed by an output layer consisting
of three neurons (one representing each fluency class). Each neuron
within the hidden layers uses the relu function as the activation
function. For the output layer, we implement the softmax function
to convert the output into class probabilities. Finally, the predicted
label is the class with the highest probability.

The convolutional neural network architecture has four hidden layers,
the first two have 64 convolution filters and the following two have
32 convolution filters. The output layer consists of three neurons
corresponding to our three classes and similarly to the multilayer
perceptron, the activation function of the hidden layers is the relu
and for the output layer is the softmax function.

The recurrent neural network architecture is a long-short-therm-memory
(LSTM) \cite{key-18}. It has two hidden layers comprising 256 x 32
neurons and three neurons corresponding to the output layer. This
architecture also implements the same activation functions as the
other two networks. 

In \textit{table 1} we summarize the architectures described above.

\begin{table}[H]
\begin{raggedright}
{\footnotesize{}}%
\begin{tabular}{|c|c|c|c|}
\hline 
{\footnotesize{}Neural Network} & {\footnotesize{}Hidden layers} & {\footnotesize{}Neurons} & {\footnotesize{}Activation }\tabularnewline
\hline 
\hline 
{\footnotesize{}MLP} & {\footnotesize{}2} & {\footnotesize{}512x512x3} & {\footnotesize{}relu, softmax}\tabularnewline
\hline 
{\footnotesize{}CNN} & {\footnotesize{}4} & {\footnotesize{}64x64x32x32x3} & {\footnotesize{}relu, softmax}\tabularnewline
\hline 
{\footnotesize{}RNN} & {\footnotesize{}2} & {\footnotesize{}256x32x3} & {\footnotesize{}relu, softmax}\tabularnewline
\hline 
\end{tabular}
\par\end{raggedright}{\footnotesize \par}
\caption{Neural networks architectures.}

\end{table}

The other two models are traditional machine learning models. One
is a support vector machine. This model has a basic construction (similar
to the one proposed in the scikit-learn documentation). The main hyper-parameter
of the estimator that we varied was the \textit{regularization parameter
C}. \footnote{\href{http://scikit-learn.org/stable/modules/svm.html}{Link to SVM scikit-learn documentation.}}

The other model uses a random forest classifier. This model also has
a basic construction in the sense that our python script only initializes
the model, trains it and then evaluates its performance. The single
parameter we varied here was the \textit{number of estimators} (number
of trees).\footnote{\href{http://scikit-learn.org/stable/modules/generated/sklearn.ensemble.RandomForestClassifier.html}{Link to RF scikit-learn documentation.}}

\section*{\textcolor{black}{\large{}4. Experimental Results}}

In this section we present our feature extraction and classification
results. The main data analysis tools that we have used for our experiments
are Anaconda (under Python 2.7), Keras (backend Tensorflow), Scikit-learn,
Librosa, Pydub and Pandas. 

We have evaluated different model constructions, here, we report the
model architectures (from \textit{Table 1}) and hyper-parameters that
achieved the highest classification accuracies. 

As a first step, we explore the effect of varying the number of Mel
coefficients ($N_{mel}$) on the final accuracy. Increasing the value
$N_{mel}$ increases the complexity of a model, thus, it is our duty
to find out the trade-off between the appropriate $N_{mel}$ and the
maximum achievable accuracy. There is a point in which adding more
$N_{mel}$ doesn't translate into considerable improvements (this
can either increase accuracy by a small percentage or decrease it).
With this experiment, we are able to get the appropriate $N_{mel}$
for our data set.

As a second step, we show how adding features such as ZCR, RMSE and
SF to the baseline chosen $N_{mel}$ boosts accuracy in most of the
cases.

\subsection*{\textcolor{black}{\normalsize{}4.1 Classification Experiments}}

We have chosen \textbf{SVM} and \textbf{RF} as our models based on
research regarding the most appropriate ML approaches for audio classification.
Both of these models have proven to be good candidates for the classification
of sound events, such as the\textit{ ECS-50} audio set, as proposed
by Piczak \cite{key-17}. 

We are also comparing three different neural network models; the \textbf{MLP}
architecture has proven to classify accurately speech, audio and noise
audio of the \textit{MUSAN} audio set as reported by Wetzel et al.
\cite{key-5}, \textbf{CNNs} have also been used to classify the \textit{ECS-50}
audio set \cite{key-4} and have also been used to classify audio
without performing prior feature extraction, in the sense that the
network itself extracts corresponding features from the waveform sample
\cite{key-12}, lastly, we have also employed \textbf{RNNs} because
according to Huy Phan et al., this type of neural networks achieved
an accuracy of 97\% in the classification of sound scenes from the
\textit{LITIS Rouen} dataset \cite{key-19}.

All our models were randomly sampled with 70\% training data and 30\%
test data. Given the 1424 total audio frames of the Avalinguo audio
set, this corresponds to 926 audio frames for training and 498 audio
frames for testing.

In our first experiment, we have trained and evaluated accuracy performance
with increasing values of $N_{mel}$. \textit{Table 2} contains the
achieved accuracies as we set $N_{mel}=5,10,12,20$. In each case,
the accuracy improved considerably as the value $N_{mel}$ increased.
We obtained an accuracy as high as 94.39\% with the SVM for $N_{mel}=20$.
We also trained our models with $N_{mel}=30,40$ but this only increased
feature space dimensionality but not accuracy. From this experiment,
we see that the commonly number of Mel coefficients used for audio-analysis
(12 to 20) applies to our data as well.

\begin{table}[H]
\begin{centering}
\begin{tabular}{|c|c|c|c|c|}
\hline 
Model & 5  & 10  & 12  & 20 \tabularnewline
\hline 
\hline 
SVM & 86.00\%  & 89.49\%  & 92.06\%  & 94.39\% \tabularnewline
\hline 
RF & 84.80\%  & 89.00\%  & 90.42\%  & 92.29\% \tabularnewline
\hline 
MLP & 78.00\%  & 88.78\%  & 89.01\%  & 92.05\% \tabularnewline
\hline 
CNN & 80.00\%  & 85.04\%  & 87.61\%  & 93.69\% \tabularnewline
\hline 
RNN & 78.90\%  & 85.04\%  & 86.44\%  & 87.00\% \tabularnewline
\hline 
\end{tabular}
\par\end{centering}
\caption{Accuracy performance of the classification models for different $N_{mel}$
values.}

\end{table}

The second stage consists in taking the $N_{mel}=20$ as baseline
features and test extra spectral features to see the outcome. After
doing feature exploration in our runs, we have ended up adding ZCR
(as proposed in \cite{key-5}), as well as RMSE and SF (as proposed
in \cite{key-7}). This translates to a 23-dimensional feature space
(20 MFCCS + 1 ZCR + 1 RMSE + 1 SF).

The architecture of our models combined with the total final features
yields a runtime of about 1 ms per classification on a 2.4 GHz single
core CPU for the neural networks. For the SF and RF, it takes about
two seconds to train completely. 

\textit{Table 3} shows the obtained accuracies with the extra features.
Once again, the SVM achieved the highest accuracy followed by the
RF. The MLP and CNN obtained similar results, the one big difference
is that the MLP outperformed the CNN in computing time; the former
took about 1min to train completely, whereas the latter took about
5 mins to train completely. The RNN obtained the lowest accuracy and
it took about 7 mins to completely train. 

In contrast with the results from \textit{table 2} (case $N_{mel}=20$),
the results in \textit{table 3} ($N_{mel}=20$ + extra features) show
that the performance of the SVM remained equal, the CNN performance
decreased by around 1\% and the other three models increased their
accuracy. We present graphically this comparison with a bar plot in
\textit{Figure 4}.

\begin{table}[H]
\begin{centering}
{\tiny{}}%
\begin{tabular}{|c|c|c|c|c|c|}
\hline 
{\tiny{}Features} & {\tiny{}SVM} & {\tiny{}RF} & {\tiny{}MLP} & {\tiny{}CNN} & {\tiny{}RNN}\tabularnewline
\hline 
\hline 
\textbf{\tiny{}$N_{mel}$+ZCR+RMSE+SF} & {\tiny{}94.39\% } & {\tiny{}93.45\% } & {\tiny{}92.52\% } & {\tiny{}92.75\% } & {\tiny{}89.01\% }\tabularnewline
\hline 
\end{tabular}
\par\end{centering}{\tiny \par}
\caption{Accuracy performance of the classification models 20 MFCCs + extra
features.}
\end{table}

\begin{figure}[H]
\begin{centering}
\includegraphics[scale=0.27]{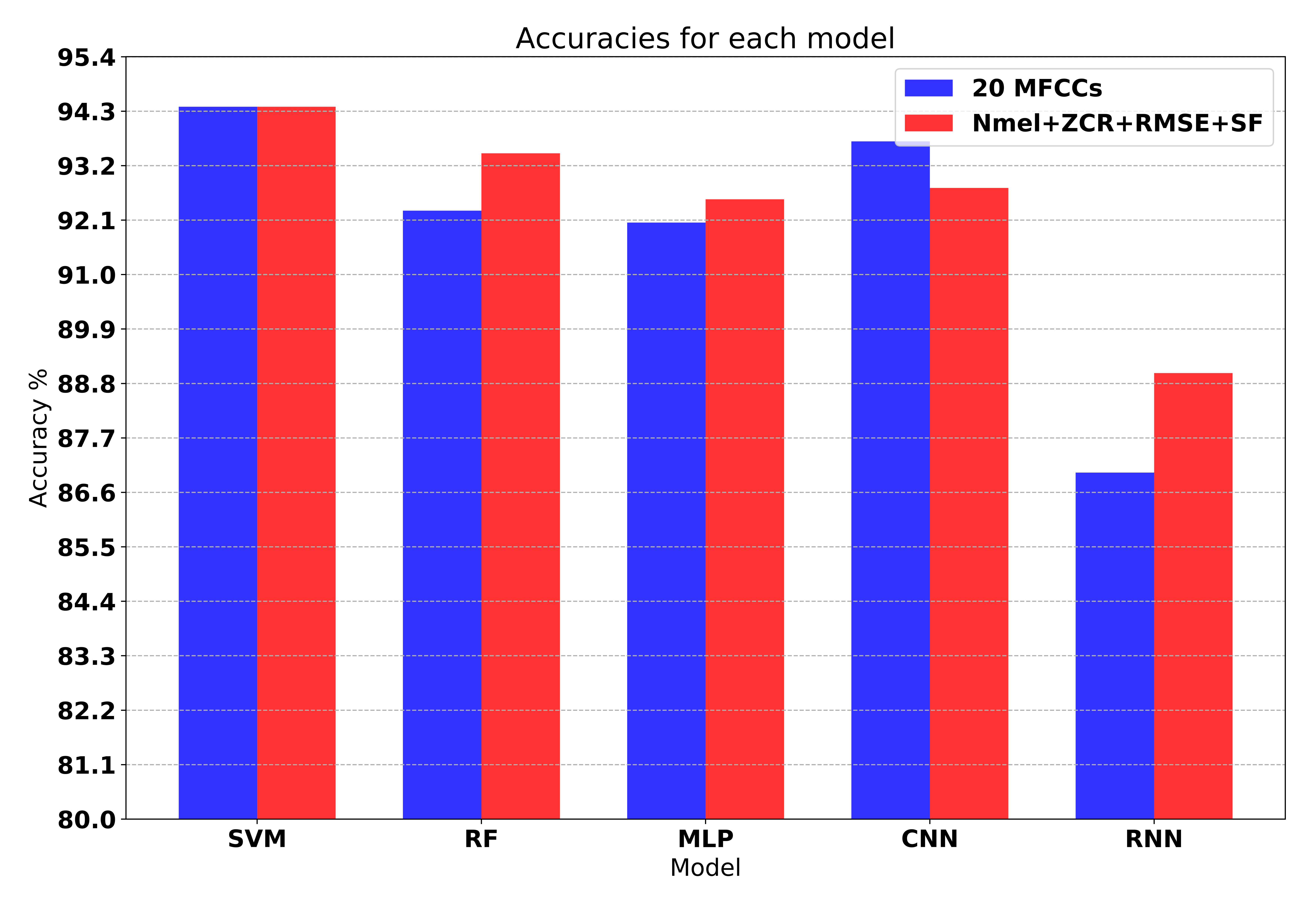}
\par\end{centering}
\centering{}\caption{Accuracy comparison when using $N_{mel}=20$ (blue bars) and with
extra features (red bars). For sake of visualization, the accuracy
of the plots starts at 80\%.}
\end{figure}

With these results, we have that the SVM followed by the RF are the
models that best classify our data set. The reason why the SVM couldn't
improve further with the extra features can have to do with the design
of the model itself. In this case, we varied the choice of the regularization
parameter \textit{C} but couldn't obtain any better performance. In
the case of the RF, we increased the \textit{number of trees} when
we added the other features, gaining more than 1\% accuracy. 

The neural networks have slightly underperformed comparing them with
the other two models. But they have achieved decent accuracies as
well. The main difference between the deep learning and the traditional
models has to do with the time it takes them to train, requiring the
latter way less computational time. The under-performance of the neural
networks does not exclude them from this analysis at all, it can be
that a different architecture can boost their accuracies. 

In order to evaluate the quality of the output of the classifiers
we have used a confusion matrix. The corresponding map can be seen
in \textit{Figure 5} and it belongs to the SVM results, which trained
to classify among our three fluency levels, has achieved a classification
accuracy of 94.39\%. The matrix was plotted using the \textit{sklearn.metrics}
module from scikit-learn \cite{key-20}. From the plot, we see that
the classifier actually predicted all the high labels (classes) correctly.
For the intermediate label, it misclassified 15 audio frames either
as a low or high label, this is understandable since, intuitively,
it is harder to discriminate if a frame lies in the intermediate level
or if it belongs to any of its ``neighbors''. For the low label
results, the SVM predicted two audio frames as highly fluent whilst
they belong to the low fluency class, this could be the ``most critical''
mistake our model has done by predicting that two audios whose fluency
is low, actually has an almost native fluency. However, in the overall,
the model has performed remarkably.

\begin{figure}[H]
\begin{centering}
\includegraphics[scale=0.35]{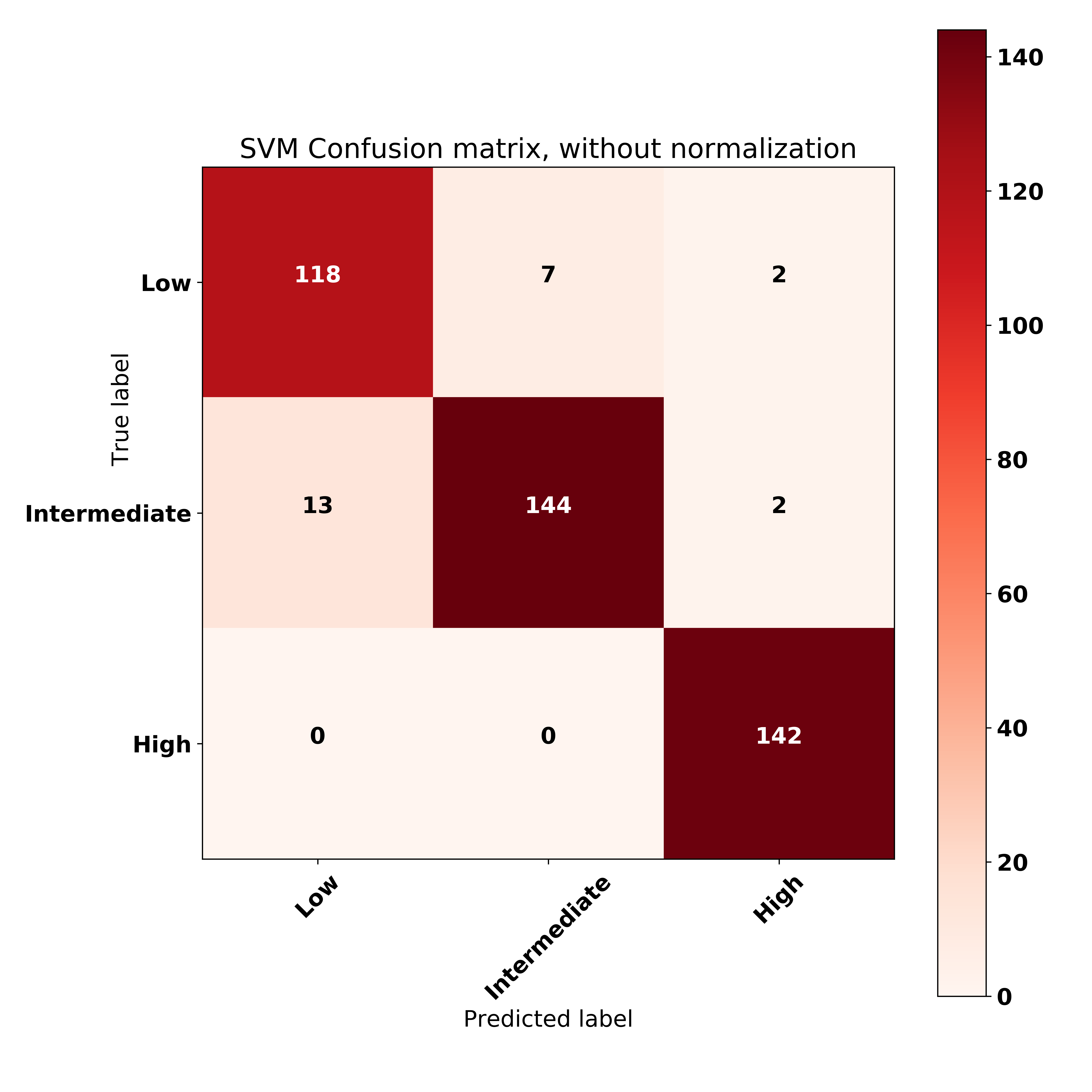}
\par\end{centering}
\centering{}\caption{Confusion matrix for the SVM trained and tested with the Avalinguo
audio set.}
\end{figure}

The link to the repository with the Python script to replicate this
paper can be found in {[}16{]}. All the technical requirements to
run the code are documented in the attached repository.

\section*{\textcolor{black}{\large{}5. Conclusions}}

In this work, we have presented an audio processing system capable
of determining the level of fluency of non-native English speakers,
taken 5s non-overlapped audio segments from the Avalinguo audio set.
We have used five different ML models to classify audio segments into
low, intermediate or high fluency levels. Each model was capable of
classifying audio frames with an accuracy of more than 90\% (except
one classifier that reached 89\%). 

As a first step, we have determined that the appropriate number of
Mel cepstral coefficients for our data set is 20. Thereafter, with
these baseline features, we have added zero-crossing rate, root mean
square energy and spectral flux features to improve the accuracy of
our models. The highest accuracies were reached by SVM and RF, with
94.39\% and 93.45\%, respectively. The neural networks achieved also
remarkable accuracies (MLP 92.52\%, CNN 92.75\%, RNN 89.01\%).

We have also reported the construction and details of the Avalinguo
audio set, whose main characteristic is that it is composed by conversations
of people who are learning the English language.

The accuracies that we have achieved can be considered high but nonetheless
there is room for improvement. For example, tunning more precisely
the hyper-parameters of the SVM and RF estimators by running grid
searches. In the case of the neural networks architecture, we can
still explore adding specific hidden layers and modifying the number
of neurons per layer. Added to this, as other works have proposed,
exploring with other audio features such as chromagram, Mel spectrograms
or spectral contrast can improve accuracy. We must take into account
that we are only defining three fluency classes, in order to make
fluency levels more specific, we would have to define more fluency
classes in between. This poses a challenge for the accuracy performance.

The main technical limitation that we have right now is that the Avalinguo
audio set (with about 2 hrs of recordings) can be considered a small
set. Part of our future work consists in maintaining and increasing
the size of the audio set. Another technical challenge consists in
automatically identifying persons within a conversation and at the
same time, capturing the silences and pauses they can make. 

This project will be integrated to the Avalinguo system. Here, we
will have to deal with the classification of live conversations, for
example.

\pagestyle{plain}

\subsubsection*{Acknowledgments}

We gratefully acknowledge the support by the Language Center at ITESM
and the individual persons who contributed with the construction of
the dataset.

\end{multicols}
\end{document}